\documentclass[twoside]{IEEEtran}
\usepackage{cite}
\usepackage{amsmath,amssymb,amsfonts}
\usepackage{algorithmic}
\usepackage{graphicx}

\usepackage{textcomp}
\usepackage{xcolor}
\usepackage{booktabs}
\usepackage{tabularx,booktabs}
\usepackage{tabularx}
\def\journalname{Under review at IEEE TBME}
\def\BibTeX{{\rm B\kern-.05em{\sc i\kern-.025em b}\kern-.08em
    T\kern-.1667em\lower.7ex\hbox{E}\kern-.125emX}}

\begin{document}

\title{Merging Deep Learning with Expert Knowledge for Seizure Onset Zone localization from rs-fMRI in Pediatric Pharmaco Resistant Epilepsy}
\author{Payal Kamboj, Ayan Banerjee, \IEEEmembership{Member, IEEE}, Sandeep K.S. Gupta, Sr., \IEEEmembership{ Member, IEEE}, and Varina L. Boerwinkle
\thanks{Manuscript submitted May 5, 2023.}
\thanks{Payal Kamboj, Ayan Banerjee and Sandeep K.S. Gupta are with the School of Computing and Augmented Intelligence, Arizona State University, Tempe, 85281 USA (e-mail: pkamboj@asu.edu, abanerj3@asu.edu and sandeep.gupta@asu.edu).}
\thanks{Varina L. Boerwinkle is with Department of Neurology, Division of Child Neurology, University of North Carolina, North Carolina, 27599 USA  (e-mail: varina\_boerwinkle@med.unc.edu).}
}

\maketitle

\begin{abstract}
Surgical disconnection of Seizure Onset Zones (SOZs) at an early age is an effective treatment for Pharmaco-Resistant Epilepsy (PRE). Pre-surgical localization of SOZs with intra-cranial EEG (iEEG) requires safe and effective depth electrode placement. Resting-state functional Magnetic Resonance Imaging (rs-fMRI) combined with signal decoupling using independent component (IC) analysis has shown promising SOZ localization capability that guides iEEG lead placement. However, SOZ ICs identification requires manual expert sorting of 100s of ICs per patient by the surgical team which limits the reproducibility and availability of this pre-surgical screening. Automated approaches for SOZ IC identification using rs-fMRI may use deep learning (DL) that encodes intricacies of brain networks from scarcely available pediatric data but has low precision, or shallow learning (SL) expert rule-based inference approaches that are incapable of encoding the full spectrum of spatial features. This paper proposes \textit{DeepXSOZ} that exploits the synergy between DL based spatial feature and SL based expert knowledge encoding to overcome performance drawbacks of these strategies applied in isolation. \textit{DeepXSOZ} is an expert-in-the-loop IC sorting technique that a) can be configured to either significantly reduce expert sorting workload or operate with high sensitivity based on expertise of the surgical team and b) can potentially enable the usage of rs-fMRI as a low cost outpatient pre-surgical screening tool. Comparison with state-of-art on 52 children with PRE shows that \textit{DeepXSOZ} achieves sensitivity of 89.79\%, precision of 93.6\% and accuracy of 84.6\%, and reduces sorting effort by 6.7 ($\pm$2.2)-fold. Knowledge level ablation studies show a pathway towards maximizing patient outcomes while optimizing the machine-expert collaboration for various scenarios. \footnote{This paper is currently under review.}
\end{abstract}

%\begin{IEEEkeywords}
%Deep Learning, Imbalanced class, rs-fMRI, Expert Knowledge,  Pharmaco Resistant Epilepsy.
%\end{IEEEkeywords}

\section{Introduction}
 \begin{figure*}[htbp]
\includegraphics[width=180mm,scale=2,height=92mm]{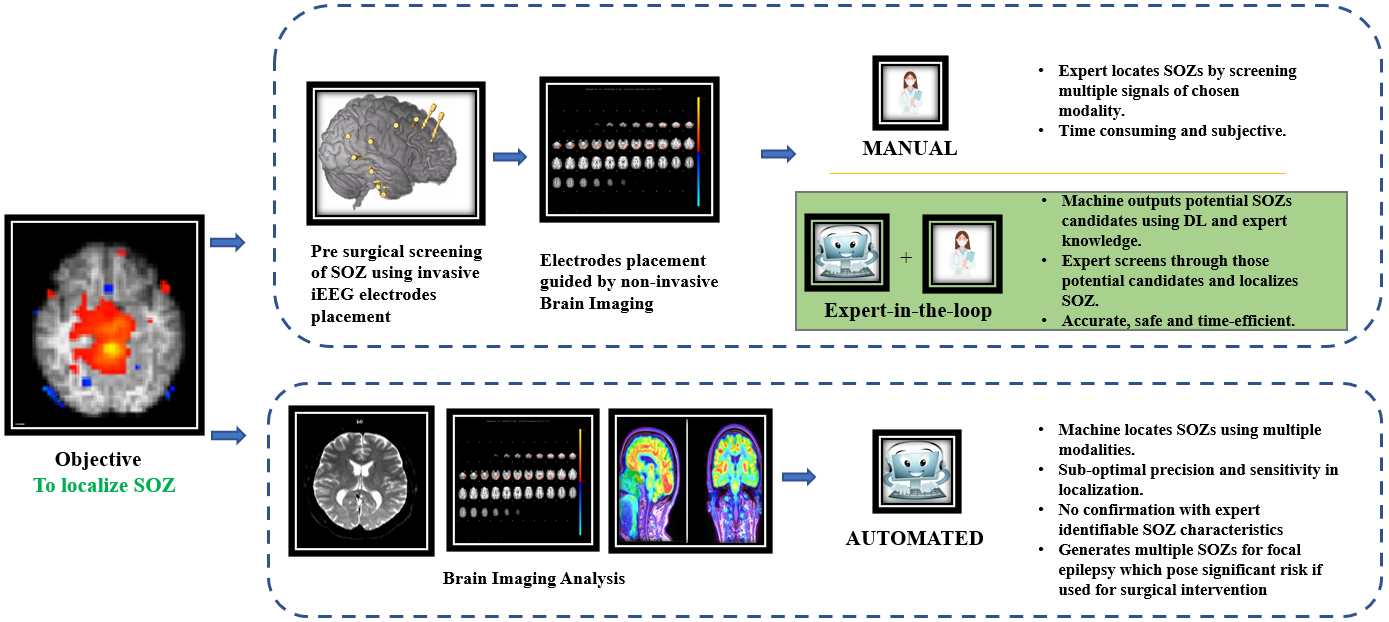}
\caption{ Non-invasive brain imaging guides the iEEG electrodes placement to localize SOZ, which can be achieved using manual, automated, or expert-in-the-loop approaches. Manual localization is subjective and time-consuming due to expert examining all relevant images/signals, while fully automated approaches may generate inaccurate or multiple SOZs. The expert-in-the-loop approach combines automated detection with expert's analysis in the end for efficient and accurate localization of the SOZ.}
\label{fig:AHM}
\end{figure*}

%\IEEEPARstart
According to the World Health Organization (WHO), around 50 million people worldwide suffer from epilepsy\cite{b12}\cite{b35}. Pharmaco Resistant Epilepsy (PRE), which accounts for 30\% of cases, occurs when a patient is not seizure free for at least 12 months through adequate trials of two tolerated and appropriately chosen anti-epileptic medications, and it immensely affects the patient's quality of life\cite{b36}. The most effective treatment for PRE is surgical resection or ablation of the Seizure Onset Zone (SOZ), the part of the brain where seizure originates~\cite{b12,b13}. Recent studies have advocated for early diagnosis and surgery to avoid developmental complications which may cause sudden deaths~\cite{b49}. In fact, surgical intervention as early as three months of age is shown to have excellent post-surgical seizure frequency outcomes (Engel scores), with negligible risk of permanent morbidity~\cite{b50}. This paper seeks to automate pre-surgical evaluation of children with PRE using rs-fMRI. The goal is to make this evaluation accessible in clinical settings, overcoming the current need for specialized centers with local expertise.

%This paper focuses on a brain imaging technique by resting state functional magnetic resonance imaging (rs-fMRI) to enable automation of accurate, time efficient, and accessible pre-surgical evaluation of children with PRE in clinical settings, that currently is limited to only centers with local expertise for interpretation.

In children, the risk of post-surgery developmental impairments and continued seizures 
necessitate precise localization of SOZ for PRE treatment success. Seizures are spatio-temporal phenomenon, which originate from a SOZ, and propagate to other parts of the brain~\cite{b47} (also known as epilepsy networks, EN)\cite{b48}. As such capturing SOZ using brain imaging is not only time sensitive but also requires high spatial resolution. Any given existing brain imaging technique does not have the required spatial and temporal resolution to automatically and accurately identify the SOZ~\cite{b44}. As a result, in pre-surgical evaluation, SOZ requires expert manual analysis of images from multi-modal sensing (Fig. \ref{fig:AHM}) such as the SISCOM method that combines ictal and inter-ictal single-photon emission computerized tomography (SPECT) scan~\cite{b53}, or a combination of Magnetoencephalography (MEG) for high temporal resolution and concordant fMRI for high spatial resolution~\cite{b30} or inter-cranial electroencephalography (iEEG) and rs-fMRI.

The gold standard technique for localization of SOZ uses iEEG, which requires implantation of depth electrodes~\cite{b51} (Fig. \ref{fig:AHM}). However, for seizure free outcomes, optimal lead placement is essential. A sub-optimal hypothesis of where to place iEEG electrodes can reduce the effectiveness of iEEG monitoring as evidenced in a study that reports good concordance between iEEG spike density and SOZ in only 56\% of patients \cite{b56}. Also, invasive methods such as iEEG come with the risk of brain injury and long term complications. Two alternatives have been explored in recent works to improve pre-surgical screening (Fig. \ref{fig:AHM}): a) guiding the iEEG lead placement to the expected SOZ location with prior rs-fMRI while avoiding functional brain networks such as motor control or memory~\cite{b11} \cite{b23}, and b) bypass iEEG monitoring with fully automated non-invasive SOZ localization using machine learning on brain images including rs-fMRI~\cite{b54}. The working hypothesis of these two approaches is driven by several engineering and clinical observations: a) fMRI is a non-invasive brain imaging technique that has high spatial resolution, b) manual analysis of rs-fMRI imaging has been shown to have 90\% agreement with iEEG determined SOZ~\cite{b51}, and c) usage of rs-fMRI in conjunction with iEEG to locate and surgically alter SOZ has shown significant improvement in surgical outcomes for children with PRE in terms of Engel scores without any increase in developmental risks~\cite{b19}.

Although adoption of rs-fMRI for SOZ identification has been limited in clinical practice, iEEG lead placement guided by manual analysis of rs-fMRI has seen application in PRE care. On the other hand, to the best of our knowledge, automation of SOZ identification using deep learning (DL) on brain imaging is a novel and untested concept in clinical settings~\cite{b12}~\cite{b16}~\cite{b24}~\cite{b44}. While these techniques are appealing, their clinical application necessitates the use of supplementary imaging techniques such as diffusion MRI (dMRI), which captures movement of molecules (mainly water) in tissues and may require additional imaging time. This can potentially increase the burden on patients. These techniques also suffer from low accuracy and sensitivity. Moreover, the DL identified SOZ may not conform with the expert identifiable SOZ characteristics. The most recent DL study that combines rs-fMRI and dMRI, localizes SOZ~\cite{b54} in both brain hemispheres for focal epilepsy subjects, which violates the basic characteristic of asymmetry of activation. This indicates that automated techniques can frequently identify normal brain locations as SOZ (in six out of 14 cases reported in ~\cite{b54}). Given the vulnerable nature of the pediatric brain, we believe that a fully automated solution to SOZ identification is currently not feasible for clinical integration. Hence, in this paper, we advocate for an expert-in-the-loop solution (Fig. \ref{fig:AHM}), where a configurable DL-based automation incorporates expert knowledge and reduces the manual effort by presenting highly relevant SOZ signals, which can be further sorted by the expert in the end with greater confidence on the SOZ localization.

%Surgical resection or ablation of SOZ identified using a combination of brain imaging and intracranial electroencephalography (iEEG) is reported to have upto 80\% \cite{b10}\cite{b13} seizure freedom success. 
%There exist several non-invasive brain imaging techniques like Positive Emission Tomography (PET),  single-photon emission computerized tomography (SPECT), and Magnetoencephalography (MEG) for SOZ localization but not only their localizing power is suboptimal\cite{b22}, it is also a time consuming process to rely on these techniques due to the infrequent nature of seizures, as one has to wait for the seizure to occur to record the data\cite{b12}. On the other hand, resting-state functional Magnetic Resonance Imaging (rs-fMRI) with Independent Component Analysis (ICA) \cite{b11} approach has shown significant efficacy in improving the surgical candidacy by accurate localization of the SOZ which as a whole improves surgical outcomes in PRE patients\cite{b7}. 
\begin{figure*}[htbp]
centerline{\includegraphics[width=170mm,height=100mm]{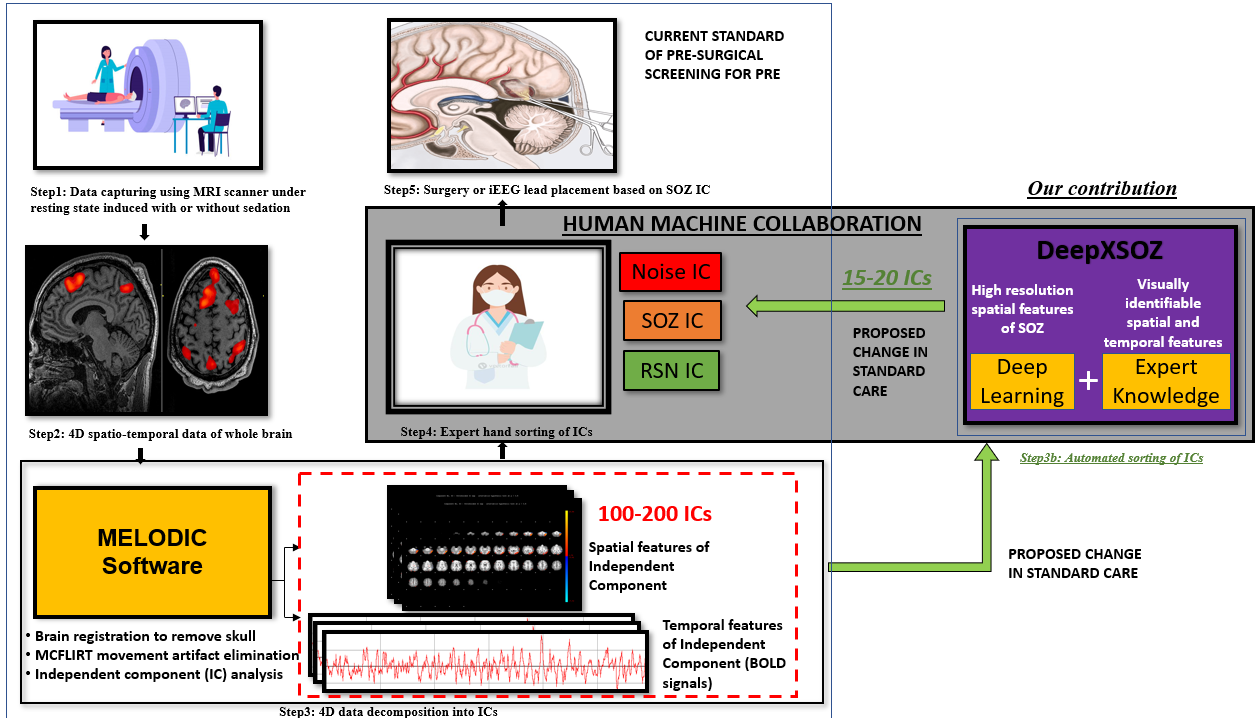}}
\caption{\textit{DeepXSOZ}'s contribution in automated sorting of ICs in PRE's pre-surgical screening.}
\label{fig:Contrib}
\end{figure*}

\subsection{Background and Problem Statement}
fMRI evaluates functional connectivity in terms of Blood Oxygen Level Dependent (BOLD) consumption activation (RED colored clusters in step 2 of Fig. \ref{fig:Contrib}). It is a composite outcome of normal brain activity, seizure activity, and noise due to head movement or measurement artifacts~\cite{b55}. To decouple the normal brain activity from seizure activity, rs-fMRI is performed on PRE patients~\cite{b42}. Independent component analysis (ICA) is a useful processing technique for fMRI that generates mutually orthogonal spatio-temporal independent components (ICs) such that each IC encodes characteristics of resting state brain activity, named as Resting State Network (RSN), seizure onset, named as SOZ, or noise~\cite{b9} \cite{b11}. 

In fMRI based pre-surgical iEEG lead placement evaluation (Fig. \ref{fig:Contrib}), the first step is to acquire rs-fMRI, which generates 4 D (3 D space and 1 D time) imaging data. The second step is pre-processing of the rs-fMRI for head motion artifact removal and ICA. MELODIC tool~\cite{b9} and MCFLIRT~\cite{b31} are popularly used softwares for this purpose. Based on the spatial and temporal resolution set at the measurement time, ICA of rs-fMRI may result in 100 - 200 ICs\cite{b17} (right hand side of Fig. \ref{fig:Contrib}'s step 3) and only a small fraction (less than 10\%) of them are SOZ ICs. Although the ICs are mutually orthogonal, ICA cannot label the ICs as RSN, SOZ or noise. At this point, expert manual sorting (Step 4) is employed, which is time-consuming, subjective, and limits the reproducibility and availability of rs-fMRI-based SOZ identification \cite{b15}. 

%expert manual sorting is employed (Step 4). This expert sorting of ICs not only requires careful observation, but it is a very time consuming and subjective process \cite{b15}. This limits the reproducibility and availability of rs-fMRI based SOZ identification. 

\begin{center}
    \textbf{\underline{Problem Statement}}
\end{center}
\noindent{\bf Given:} a) A patient with age $<= 18$ diagnosed with pharmaco resistant focal epilepsy.

\noindent b) A set of $N (\approx 100)$ rs-fMRI ICs (with both Spatial and Temporal evolution of BOLD signals) potentially consisting of three groups: i) noise ICs (55\%, ICs primarily affected by measurement noise), ii) RSN ICs (40\%, with activation primarily affected by resting state function of the brain) and iii) SOZ ICs ($<5\%$, with activation affected by seizure onset).

\noindent{\bf Automatically Derive:} A set of ICs with high likelihood that they are primarily affected by seizure onset.

%\begin{center}
%\textit{In this paper, we present DeepXSOZ (Step 3b in Fig. \ref{fig:Contrib}), which performs expert-in-the-loop automation of Step 4.} 
%\end{center}

 \subsection{Challenges of automation}
 \noindent{\bf Types of automation:} Several automated machine learning (ML) based techniques for IC sorting have been explored recently (Table \ref{LR}). Such techniques can be broadly classified into three classes: a) unsupervised techniques, b) shallow learning (SL), and c) DL. The most recent attempt, EPIK~\cite{b44}, is an unsupervised method, which has good accuracy and sensitivity, but has a significant number of FPs, where non-SOZ ICs are wrongly identified as SOZ. SL techniques suffer from significant False Negatives (FN)s, where legitimate SOZ ICs are not well identified. DL is a promising approach that has been explored in IC sorting for both healthy adults~\cite{b20} with n = 2000 and on a limited dataset of children with PRE~\cite{b44}. While DL methods show $> 90\%$ accuracy in identifying RSN, their sensitivity in identifying SOZ ICs was less than 20\%~\cite{b44}. 

%\begin{figure*}
%\centering
%\includegraphics[width=\textwidth]{sampleIC.png}
%\caption{Zoomed-in views of Noise, RSN and SOZ ICs.}
%\label{fig:sampleIC}
%\end{figure*}

\begin{figure}
\centering
\includegraphics[width=88mm,height=38mm]{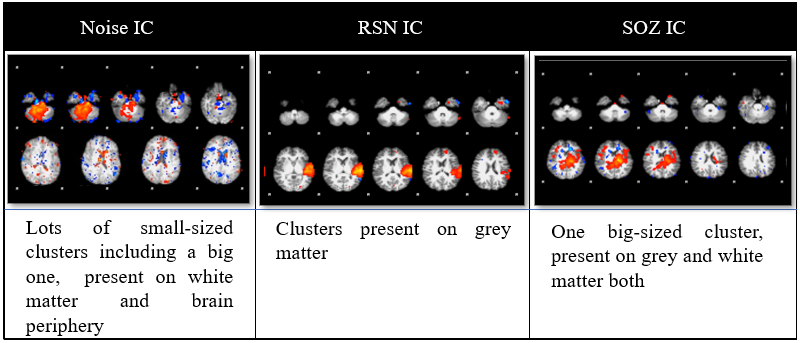}
\caption{Zoomed-in views of Noise, RSN and SOZ ICs.}
\label{fig:sampleIC}
\end{figure}

 \noindent{\bf Intra-class variability:} Automation of SOZ localization is challenging not only due to its extremely limited data availability but also because ICs suffer from large intra-class variability that cannot be addressed easily even with existing DL techniques\cite{b34}.  Moreover, DL techniques maybe biased towards the majority class and as a result report a high number of True Negatives (TNs) which further reflect high accuracy and specificity \cite{b34}. This is misconstrued  as a good result when the numbers are only higher due to the presence of a high number of TNs. Also, Noise and SOZ ICs can be visually challenging to distinguish for non-experts. Leveraging expert knowledge can help incorporate high-level SOZ properties into IC sorting, reducing variability within classes (Fig.\ref{fig:sampleIC}). 

 %The zoomed-in views of Noise, RSN and SOZ IC also demonstrate this in Figure \ref{fig:sampleIC}.

\begin{table*}
\caption{Review of fMRI based IC sorting ( Not Applicable (NA), Accuracy (Acc), Sensitivity
(Sens), Specificity (Spec), Precision (Prec), Not Specified
(NS), Epilepsy networks (EN), Epileptogenic zone (EZ), CD - cannot deduce, Patient level metric (PLM, Section \ref{sec:PLM}) , IC level metrics (ILM, Section \ref{sec:ILM}) ), M in the study column indicates manual, A indicates automation.}
\begin{center}
\scriptsize
\begin{tabular}{|p{0.4 in}|p{0.65 in}|p{0.15 in}|p{0.15 in}|p{0.65 in}|p{0.45 in}|p{1.7 in}|p{1.55 in}|}
\hline
\textbf{Problem} & \textbf{Study } & \textbf{N}  & \textbf{PRE} & \textbf{Age (years)}&\textbf{IC Class}  & \textbf{Reported results} & \textbf{Deduced results}\\ 
          \hline
 Epilepsy  &Nyugen\cite{b39} A  & 322         
         &63         &Child (4 - 25)  & NA  & Sens= 85\%, Acc = 71\%, 
Spec = 71\% & NA\\ 
    \cline{2-8} 
     Detection &Lopes\cite{b40} A & 15         & 0  & Adult ($>$ 18)      & NA         & Acc = 87.5\% &NA\\ 
     \cline{2-8}
  &Bharath\cite{b38} A & 132        & 0 & Adult ($>$ 18)        & EN         & Sens = 100\%, Acc = 97.5\%,  Spec = 94.4\% &NA
 \\ 
    \hline
 SOZ &Boerwinkle \cite{b19} M & 40       &40  &Child (1.5 - 19.8) & EZ-SOZ               & Agreement with iEEG derived SOZ = 90\%,  Prec = 79\%, Sens = 93\% & PLM: Prec: 79\%,  Sens: 93\%, Acc 90\%,  Spec: CD; ILM CD \\
   \cline{2-8}
   Localization &Gil\cite{b25} M &21         &0 &Adult ($>$ 18) &SOZ                 &NS &PLM: CD; ILM: CD
 \\
  \cline{2-8}
  &Lee\cite{b45} M &29        &29 &Adult ($>$ 18) &SOZ                  &Concordance with iEEG derived SOZ = 72 \% & PLM: Prec = 76\%, Sens = 76\%, Acc = 72\%,  Spec = 66\%; ILM: CD\\
  \cline{2-8}
   &Hunyadi\cite{b12} A & 18         &18   & Adult ($>$ 18)       & SOZ         & PLM: Sens=40\%, Acc=51\%, Spec=77\%&PLM, ILM reported in this study \\ 
     \cline{2-8}
     &Nozais\cite{b20} A & 2093                 &0 & Adult ($>$ 18) & RSN       & Acc = 92\% & PLM NA; ILM NA\\ 
   \cline{2-8}
  %  &Varatharajah \cite{b13} & 82       &82 &A  & SOZ                & Acc = 74.22\% &\\
  % \cline{2-8}
   &Luckett \cite{b24} A &2164        &0 &Adult ($>$ 18) &SOZ                & Lateralization of epilepsy foci Acc = 90 \% as compared to video EEG  & PLM: NA; ILM: NA\\
   \cline{2-8}
 
    \cline{2-8}
    &Zhang\cite{b16} A  & 10        &10 & Adult ($>$ 18)  & SOZ  & Consistency with physicians assessment & PLM:  Prec = 70\%, Sens =77\%, Acc = 41\%, Spec = 57\% ; ILM CD  \\ 
    \cline{2-8}
    &Banerjee\cite{b44} A       &52         &52  &Child (0.25 - 18) &RSN, SOZ                & PLM: Prec = 93\%, Sens = 79\%, Acc = 75\% & ILM reported in this paper
 \\
    \cline{2-8}    
   &\textbf{Kamboj} A &\textbf{52}         &\textbf{52}         &Child (0.25 - 18) &\textbf{RSN, SOZ}        & PLM: Prec = 93\%, Sens = 89\%, Acc = 84\%  &  ILM reported in this paper\\
   \hline
  \end{tabular}
  \label{LR}
\end{center}
\end{table*}

\noindent{\bf Data Imbalance:} DL techniques have poor performance in applications with imbalanced data as shown in many different domains including fMRI image analysis~\cite{b28}\cite{b34}. In medical imaging, the image capture protocols, as well as the psycho-physiological processes affecting the imaging outcome may change with several covariates such as age, sex, individual and family disease history, prior medical intervention and other co-occurring medical conditions~\cite{b52}. In such scenarios, DL performance may be negatively affected due to the lack of pathological data for a specific group of subjects, especially the pediatric subgroup. To address such issues, expert knowledge can help to uniquely identify the characteristics of the imbalanced class (SOZ IC). DL techniques can be augmented with automated extraction of such knowledge with fast SL techniques to achieve highly accurate localization of SOZ. 

%\textit{This paper presents DeepXSOZ (Step 3b box in Fig. \ref{fig:Contrib} ) that utilizes the synergistic benefits of combining data driven DL techniques with expert knowledge guided shallow learning in the determination of SOZ ICs for children with PRE.}.

%In this paper, we evaluate the performance of \textit{DeepXSOZ} and other contemporary techniques with respect to two goals: a) \textbf{statistical evaluation} of \textit{DeepXSOZ} in improving effectiveness of the pre-surgical screening of pediatric PRE population, and b) reducing manual IC sorting effort of the surgical team by \textbf{optimizing machine-expert collaboration}. To this effect, this paper uses \textit{patient level metrics}, for the first evaluation goal, and \textit{IC level metrics}, for the second goal. 
\subsection{Solution Approach}
This paper presents \textit{DeepXSOZ} (Step 3b box in Fig. \ref{fig:Contrib} ) that utilizes the synergistic benefits of combining data driven DL techniques with expert knowledge guided SL in the determination of SOZ ICs for children with PRE.

\subsubsection{Expert knowledge on SOZ characteristics}
\label{Expertknowledge}
Hunyadi et al.\cite{b12} and Boerwinkle et al.\cite{b19} have reported several high level characteristics of the manifestation of an SOZ in rs-fMRI IC images summarized as follows:\\
\noindent{\bf Number of clusters:} SOZ IC ideally has one cluster whereas RSN IC consists of multiple clusters (Fig. \ref{fig:sampleIC} rightmost panel).

\noindent{\bf Activation extended to ventricles:} SOZ has activation extended to ventricles through the white matter (Fig. \ref{fig:sampleIC}). 

%\noindent{\bf Minimal white matter overlap:} SOZ activation should minimally potrude into white matter if not extended to ventricles.

%Largest cluster overlapping on the grey matter indicates more likely it is to be SOZ.\\
\noindent{\bf Dominant frequencies:} SOZ BOLD signal power spectra exhibit dominant frequencies greater than 0.073 Hz.

\noindent{\bf Irregular patterns:} The rs-fMRI SOZ should demonstrate power spectra at frequencies higher than those found in RSN, and the BOLD time series may display irregular patterns.
%\noindent{\bf Sparsity in activelet basis:} Activelets are a dictionary of wavelet basis functions which fit the BOLD signals. SOZ signal is made up of sparse transient events and thus has a sparse representation in the activelet basis.

%\noindent{\bf Sparsity in sine dictionary:} Time courses of RSN are characterized by low frequency (0.01-0.1Hz) signals and thus show a sparse representation in this frequency band.

%\subsection{Challenges of applying DL in SOZ IC Identification}
%\begin{itemize}
%\item SOZ localization with DL is challenging due to the limited data availability. The rs-fMRI dataset also has large intra-class variability that cannot be addressed with existing DL techniques\cite{b34}. Moreover, DL techniques are biased towards the majority class in such cases \cite{b34}. 
%\item ICs, especially Noise and SOZ have an overlap of features that are difficult for non-experts to visually to differentiate. Thus, the only current way to automate this classification is to include the domain knowledge in the process. Fig. \ref{fig:sampleIC} shows examples Noise, RSN and SOZ IC for demonstration.
%\end{itemize}

\subsubsection{Summary of Contributions}
Fig. \ref{fig:Contrib}, shows our automation objective in the standard rs-fMRI based pre-surgical screening workflow. The primary goal is to minimize the number of ICs for the surgical team to evaluate in iEEG lead placement or surgical procedures. The main contributions include:
\begin{itemize}
\item A novel synergistic algorithm \textit{DeepXSOZ} that combines DL classification results with expert knowledge on SOZ characteristics through SL to achieve automated identification of SOZ localizing ICs which are relatively infrequent in the dataset. 
\item \textit{DeepXSOZ} partially automates Step 4 in this process and significantly reduces manual sorting effort ( 7-fold ) from the neurosurgical team. %It achieves areduction in the number of ICs to be analyzed.
\item Comparison of \textit{DeepXSOZ} with state of art SL technique LS-SVM, CNN based DL technique and EPIK is presented on 52 children with PRE stratified across age, sex, and one-year post-operative Engel outcomes for rs-fMRI guided resection or ablation.
\item We evaluate the performance of \textit{DeepXSOZ} and other contemporary techniques with respect to two goals: a) \textbf{statistical evaluation} of \textit{DeepXSOZ} in improving effectiveness of the pre-surgical screening, and b) reducing manual IC sorting effort of the surgical team by \textbf{optimizing machine-expert collaboration}. To this effect, this paper uses \textit{patient level metrics}, for the first evaluation goal, and \textit{IC level metrics}, for the second goal.   
\end{itemize}

%The paper is organized as follows: Related work and data collection section gives an overview of the literature survey done in this area and describes the patient dataset collection process. In the methodology section, we give an overview of our proposed methodology and detailed explanation of three steps of our technique \textit{DeepXSOZ}. In the end, we discuss our findings in the results and discussion section. 

\section{Related Works}
Recent works broadly fall into two categories (Table \ref{LR}): \textbf{Epilepsy detection} \cite{b38}\cite{b39}\cite{b40}, which involves classification of patients as epileptic or non-epileptic based on EN identification, and \textbf{SOZ localization}\cite{b12}\cite{b22}\cite{b41}, which is the main focus of this paper. Table \ref{LR} provides a comparative analysis of recent works in terms of number of subjects, PRE subgroup proportion, age range, and the types of ICs that are identified. Works in this domain include varied evaluation metrics such as concordance with iEEG, agreement with expert identified SOZ, and consistency with physician assessment. In Table \ref{LR}, the reported results column provides the evaluation metrics in the original manuscript for each work. We have recovered the meta-data for each paper (whenever available or applicable) and determined the patient level metrics, and IC level metrics defined in Section \ref{sec:ILM} and \ref{sec:PLM}, which is reported in the deduced results column. 

\noindent{\bf Manual SOZ Identification using rs-fMRI ICs:} Several manual SOZ identification techniques describe expert-developed rules on SOZ specific spatio-temporal characteristics of BOLD signals captured by rs-fMRI. Boerwinkle et al.\cite{b19} investigated the agreement between epileptogenic zone (EZ) by rs-fMRI and SOZ located by using iEEG data with prevalence-adjusted bias adjusted kappa on 40 patients and found the concordance to be 89\%. This paper revealed the weakness of previous techniques of using the most abnormal anatomical region of the brain to localize SOZ. Gil \cite{b25} manually studied 21 patients with extratemporal focal epilepsy to identify SOZ related ICs in fMRI data using the general linear model-derived EEG-fMRI time courses associated with epileptic activity. \textbf{Their primary findings suggested the ICA guided by their technique has the potential to identify epileptic network in patients with focal epilepsy}. Lee et al.\cite{b45} also manually investigated the functional connectivity changes in the ENs from rs-fMRI data using intrinsic connectivity contrast (ICC) to evaluate the non-invasive pre-surgical diagnostic potential for SOZ localization. The agreement of fMRI-IC with intracranial EEG SOZ was 72.4\%.

\noindent{\bf SL approach for SOZ identification using rs-fMRI ICs:} The first automation attempts were from Hunyadi et al.\cite{b12}, who present a set of SOZ spatial and temporal features used to train a Least-Squares Support Vector Machine (LS-SVM). The technique required several manually executed substeps and evaluation on 18 pediatric PRE patients showed high false negatives (FN). DL was first explored by Nozais et al.\cite{b20} to classify RSN vs noise ICs on non-PRE patients and reported an accuracy of 92\%. However, they did not pursue SOZ identification. Zhang et al.\cite{b16} proposed ICA based automated method using unsupervised algorithm (UA) to localize the SOZ. SOZ ICs were screened based on peripheral noise IC removal, asymmetry and temporal features (excluding IC outside of frequency band 0.01-0.1hz). Consistency with the resection surgery on 10 patients was reported. If we assume consistency as true positive (TP), failure as FN and success in rejecting non-SOZ IC as true negative (TN) and failure to reject non-SOZ ICs as false positive (FP) then the results indicate significant FPs. No other work is reported on automation of expert sorted ICA-based SOZ classification (as highlighted by Nandakumar et al~\cite{b54} and Banerjee et al~\cite{b44}) except EPIK~\cite{b44}. Both Banerjee et al~\cite{b44} and Nandakumar et al~\cite{b54} show that cursory automation attempts on prior implementations of ICA based SOZ identification such as Hunyadi et al~\cite{b12} and Zhang et al~\cite{b16} were inadequate. Banerjee et al~\cite{b44} proposed EPIK, which utilized state-of-art image processing techniques to codify expert rules gathered from multiple sources (Boerwinkle et al.\cite{b19} and  Hunyadi et al.\cite{b12}) and applied them in a carefully crafted waterfall technique (explained in detail in Section \ref{sec:Comp}). This technique reports high accuracy but poor precision.

\noindent{\bf DL approach for SOZ identification using rs-fMRI without ICA:} DL techniques have recently be employed to detect SOZ using fMRI data. Luckett et al.\cite{b24} used 2132 healthy control data for training of 3D CNN and tested it on temporal lobe epilepsy to determine SOZ was lateralization, i.e. identifying brain hemisphere with SOZ. The training data was synthetically altered in randomly lateralized regions which helped in detection of biological SOZ's hemisphere. Note that \textbf{ICs were not used here}, so this work detected the whole brain hemisphere of seizure onset rather than the brain region pointing towards the SOZ. Nandakumar et al. ~\cite{b54} explored deep graph neural networks using the T1 weighted images from rs-fMRI along with dMRI measurements. Study on 14 subjects showed a sensitivity of 40\% and precision of 52\% while an accuracy of 88\%. This discrepancy in accuracy and precision is indicative of the problem of DL techniques with imbalanced data, where the unique characteristics of the rare class (SOZ in this case) are not captured due to lack of training data. The work also highlights another drawback of DL techniques, where the brain regions identified as SOZ do not conform with the expert knowledge about SOZs. For example, in 5 out of 14 patients with focal epilepsy (rows 2, 5, 11, 12, and 13 in Fig 4 in~\cite{b54}), SOZs were identified in both hemispheres, while the true SOZ was only in one hemisphere.

\textit{DeepXSOZ} attempts to exploit the best of both the worlds, the capability of DL to extract spatial features and the capability of SL to encode expert knowledge.
%This literature review has been summarized in Table \ref{LR}. Our work addresses the challenge of SOZ localization and not epilepsy detection. Some of the abbreviations used in this table are: Not Applicable (NA), Accuracy (Acc), Sensitivity (Sens), Specificity (Spec), Precision (Prec) and Not Specified (NS).

\section{Methodology}

\subsection{Data Collection}
Retrospective analysis for this project was approved by the Phoenix Children Hospital (PCH), Institutional Review Board (IRB 20-358). We extracted a rs-fMRI dataset of 52 children with PRE from the PCH clinical database. 
\begin{figure*}[t]
\centering
\includegraphics[width=153mm,scale=1.9,height=86mm]{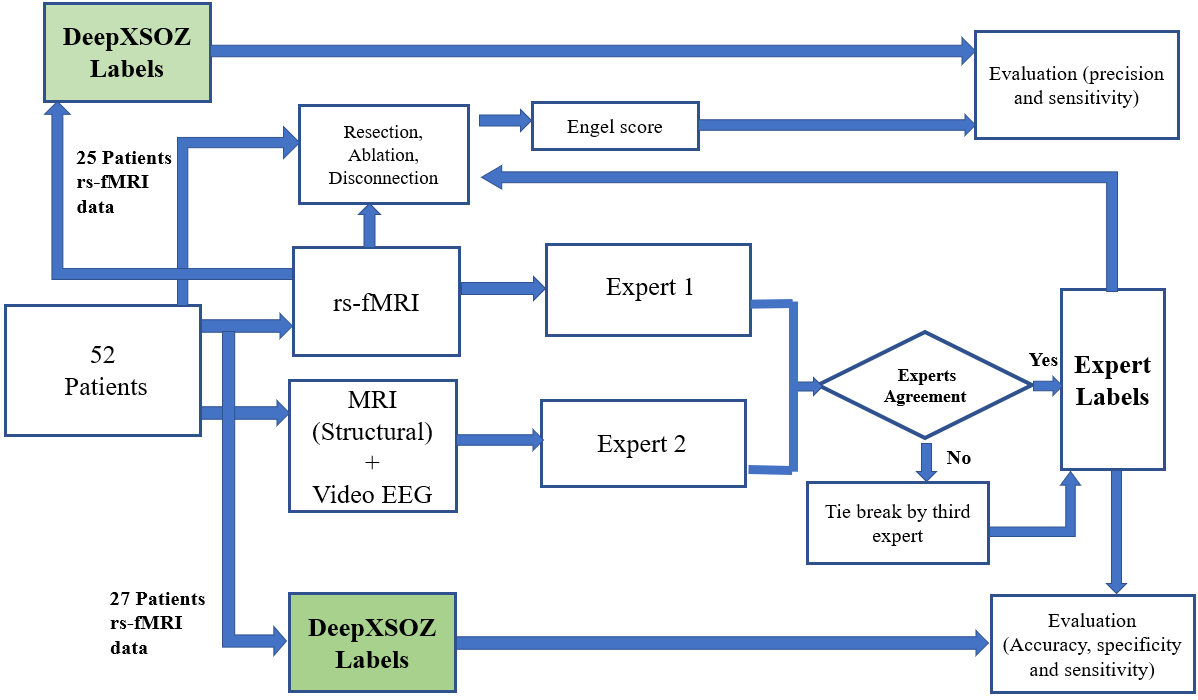}
\caption{SOZ evaluation and validation architecture using rs-fMRI.}
\label{fig:dataEval}
\end{figure*}
The data collection was performed by the PCH team without any knowledge of the \textit{DeepXSOZ} algorithm. 

\subsubsection{Inclusion Criteria}
Patients who were determined to be PRE by a treating epileptologist and received surgery evaluation. Most of the patients had focal epilepsy, however, rapid generalization of epileptiform activity from an epileptogenic focus may appear to be generalized epilepsy when evaluated using surface EEG. Hence generalized epilepsy was not an exclusion criterion for pre-surgical evaluation.
\begin{table}[htbp]
\caption{Patients Distribution.}
\scriptsize
\begin{center}
\begin{tabular}{|c|c|}
\hline
\textbf{Number of subjects}& \textit{ 52}  \\ 
    \hline
    \textbf{Age $<$ = 5 years} & 20                \\ 
     \hline
   
     \textbf{Age $>$ 5 and $<$ = 13 years}  & 18                \\ 
     \hline
   \textbf{ Age $>$ 13 and $<$ = 18 years} & 14               \\ 
    \hline
   \textbf{ Male/Female} & 23/29              \\
   \hline
   \textbf{Prior surgery} & 2             \\
   \hline
\textbf{Surgery post resting state fMRI} & Ablation 15, resection 7            \\
    \hline
   \textbf{Engel 1 score} &Ablation 10, resection 6\\  
    \hline
    \textbf{Engel 2 score} &Ablation 5 , resection 1 \\  
     \hline
     \textbf{Engel 3 score} &Ablation 1 , resection 0\\  
    \hline 
  \end{tabular}
  \label{demographics}
\end{center}
\end{table}
\subsubsection{Resting state fMRI collection method}
The rs-fMRI PCH pediatric DRE dataset from 52 children with PRE, age 3 months – 18 years old, were selected in descending alphabetical order of patient ID, who were under the care of a treating epileptologist at PCH (Table \ref{demographics}). The diagnosis of PRE was according to the treating epileptologist’s documented medical record notes. The children received rs-fMRI, video EEG and anatomical MRI as part of standard clinical MRI SOZ localization for epilepsy surgery evaluation. For rs-fMRI, patients who were determined to require conscious sedation received a propofol infusion, as a part of standard care determined by the institution’s policies. In the 52 children, 41 required conscious sedation. The dataset comprised patients with minimal head motion (less than 1 mm) during scanning.
The MRI images were acquired using a 3T MRI, Ingenuity Philips Medical systems with a 32 channel head coil. The rs-fMRI parameters were set at TR 2000 ms, TE 30 ms, matrix size $80 \times 80$, flip angle 80°, number of slices 46, slice thickness 3.4 mm with no gap, in-plane resolution $3 \times 3$ mm, interleaved acquisition, and number of total volumes 600, in two 10 -min runs, with total time of 20 mins. 
%The dataset included only the patients who had less than 1 mm head motion in any direction during scanning. 

\subsubsection{rs-fMRI pre-processing}
Oxford Center FMRIB (Functional MRI of the Brain) Software Library tool MELODIC~\cite{b9}, was used to analyze the rs-fMRI and extract ICs ~\cite{b30}. Pre-processing included deletion of the first 5 volumes to remove T1 saturation effects, passing through a high-pass filter at 100 seconds, slice time correction, spatial smoothing of 1-mm full-width at half maximum, and motion corrected by MCFLIRT~\cite{b31}, with nonbrain structures removed. Linear registration was performed between the individual functional scans and the patient’s high-resolution anatomical scan\cite{b32} which was further optimized using boundary-based registration~\cite{b33}. Individual rs-fMRI datasets then underwent ICA~\cite{b19}.

\subsubsection{RS-fMRI ground truth determination methodology}
The ground truth determination methodology should provide us with the true labels of each rs-fMRI IC as noise, RSN or SOZ. To obtain the ground truth, for each subject in addition to rs-fMRI, video EEG, and anatomical MRI data were also collected. These three modalities were independently reviewed by two blinded experts (Fig. \ref{fig:dataEval}), a neurologist, and a neurosurgeon, and each sorted and labeled the ICs into three categories - noise, RSN, and rs-fMRI SOZ. In cases where there was any disagreement, a third reviewer was consulted for the final determination. 

For patients that did undergo surgery, the surgical location was determined by the expert epilepsy surgery conference team using the expert identified rs-fMRI based SOZ location following the protocol described in Fig. \ref{fig:dataEval}. The Engel I and II scores one year after the surgery serve as the ground truth for the surgical patients.

\subsection{DeepXSOZ Architecture}

\begin{figure*}[t]
\centering
\includegraphics[width=182mm,scale=1.9,height=73mm]{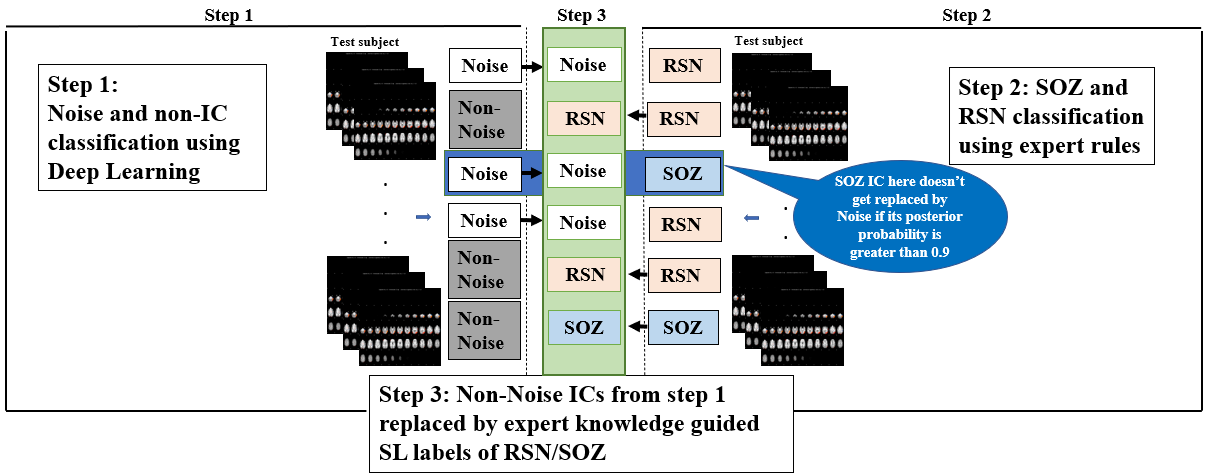}
\caption{Overview of \textit{DeepXSOZ} Architecture.}
\label{fig:DeepSOZ}
\end{figure*}

\textit{DeepXSOZ} is a three-step process (Fig. \ref{fig:DeepSOZ}).  Step 1 trains a DL network to classify ICs into noise and non-noise (RSN and SOZ ICs). The goal of Step 1 is to specifically learn to identify the Noise ICs. Step 2 further extracts features of RSN and SOZ ICs using expert knowledge and generates synthetic features of SOZ using Synthetic Minority Oversampling Technique (SMOTE). The ICs are then trained using a linear-SVM. Note that we do not use noise ICs for training in Step 2. In the final step, test subject's ICs are classified using both Step 1 and Step 2's trained machines to update the IC labels as Noise, RSN, or SOZ.

\subsubsection{Step 1: Noise ICs detection using DL}\label{AA}
ICA decomposes the 4D data into spatial and temporal ICs, where the spatial ICs are 2D images. To classify these ICs as Non-noise (RSN/SOZ) and noise ICs, 2D CNN was chosen due to its excellent performance in image classification tasks. The CNN is particularly well-suited for image analysis because it can automatically learn spatial features that are useful for classification. The hyperparameters of the CNN were tuned using Keras-tuner's hyperband algorithm with the least validation loss objective: number of convolutional layers: [3; 4; 5], number of units/filters per convolutional layer: minimum = 32, maximum = 512, default = 128, number of neurons in dense layer: minimum = 192, maximum = 1024, step = 256, learning rate: [0.01; 0.001; 0.0001], dropout rate: [0.2; 0.33; 0.4; 0.5; 0.66]. We used an image data generator by Keras to generate batches of noise and non-noise IC images, and provided real-time normalization of the images by rescaling them. We did a validation split of 0.1. Using the flow from directory method, we resized the IC images from 1006 × 709 × 3 to 270 × 400 × 3 for faster computation and used the 'binary' class\_mode and 'rgb' 'color\_mode'. 'Binary cross-entropy' was used as a loss function, and 'Adam' was used as an optimizer. To avoid the overfitting problem, regularization method called “dropout”, and “early\_stopping” strategies were implemented. “ReLU” being more computationally efficient was used as an activation function for the input and hidden layers, and “Sigmoid” activation function was used for the output layer. For CNN, weights were initialized using the “He uniform” initializer. As our dataset's images background is dark and we were required to extract the sharp features as well as reduce the variance and computation complexity, we used a max pooling layer of 2 × 2 after every convolutional layer. The optimized hyperparameter values from Keras-tuner were: Number of convolutional layers: 3, number of 3 × 3 filters in convolutional layer 1, 2 and 3: 64, 64 and 256 respectively, number of neurons in the dense fully connected layer: 704, learning rate: 0.0001 and dropout rate of 0.33. The Step 1's code was written in Python 3.7. 

\subsubsection{Step2: Feature Extraction using Expert Knowledge}
In this method, we used the SOZ specific expert rules explained by Hunyadi et al.\cite{b12} and Boerwinkle et al.\cite{b19} for feature extraction, also specified in subsection\ref{Expertknowledge}. This step is further categorized in four parts.\\
%Step 2 (Fig. \ref{fig:Step2}) works on only RSN and SOZ ICs. This step is further categorized in four parts. \\
% \begin{itemize}
%\item[a)]
\noindent {\bf a) Slice extraction}: This part extracts the brain slices from RSN and SOZ ICs, enabling us to further extract expert-guided features from these specific slices.

%\item[b)] 
\noindent{\bf b) Expert features extraction}: For each slice, we estimate the number of clusters using density-based spatial clustering of applications with noise (DBSCAN)\cite{b37}. This approach involves two adjustable parameters: neighborhood, which defines the distance metric and a value called $\epsilon$, and vmin, which determines the minimum number of neighboring voxels. Voxels with more than vmin neighbors within the $\epsilon$ distance are considered core points and form a cluster. Voxels that are not core points but are within $\epsilon$ distance of a core point are classified as border points and assigned to the nearest core point's cluster. All other points are disregarded. Clusters are formed by combining core points that are within $\epsilon$ distance of each other. To identify the activation of the SOZ that extends into the ventricles through the white matter, we employed a sobel filter-based edge detection technique \cite{b44} which allowed us to extract the contours for each slice, with the white matter exhibiting the most prominent contour within the slice. To obtain the ventricular regions, we applied the edge detection to determine brain boundary and then selected slices which exhibit more than one brain boundary contour as given by the sobel filter. The ventricular regions are within the convex hull of the brain boundary contours but do not intersect any brain boundary. Subsequently, an analysis was conducted to determine the number of larger clusters (with a size exceeding 135 pixels) that overlapped with the white matter and extended towards the ventricles. For temporal SOZ characteristics, ICs were analyzed for activelet and sine dictionary sparsity in their time courses. For calculating the sparsity in activelet basis, the BOLD signal was divided into windows of length 256 samples. From every window, four levels of activelet transformation coefficients using the 'a trous' algorithm with exponential-spline wavelets were extracted. The Gini Index metric was used for activelet coefficients and sine dictionary sparsity evaluation in the frequency band of 0.01Hz to 0.1Hz. 

%\item[c)] 
\noindent{\bf c) Balanced Dataset creation}: We generate synthetic SOZ features using SMOTE.  Since the number of SOZ ICs available is limited, approximately 6 SOZ ICs per subject, SMOTE selects the real SOZ ICs samples in the feature space, and linearly interpolates features \cite{b29}. 
%\begin{itemize}

%\item[d)]
\noindent{\bf d) SL for classification}: We employed a SVM with both linear and Radial Basis Function kernels to train a classifier using expert features extracted from RSN and SOZ. The classifier's performance was significantly superior with linear kernel. To quantify the classifier's accuracy, we used k (52) -fold cross validation. Every patient got tested using the k-1 training datasets. Similarly, we also performed k-fold cross validation for every patient using CNN to get the noise and non-noise ICs in step 1. The step 2's classifier code was written with MATLAB R2021b on the same machine as Step 1.
% \end{itemize}

\subsubsection{Step3: Combining the outputs of step 1 and step 2}
In this step, fig. \ref{fig:DeepSOZ}, we follow a specific procedure. Firstly, we subject the test subject's ICs to the machine trained in Step 1, which categorizes the ICs as either noise or non-noise. Subsequently, we pass the same test subject through the machine trained in Step 2, which assigns labels to the ICs as either SOZ or RSN. Utilizing the label list generated in Step 2 as our reference list, we apply the following criteria: if an IC is classified as noise by Step 1 but classified as SOZ by Step 2 with a posterior probability \cite{b46} less than 0.9, it is replaced with the label noise. However, if the posterior probability is greater than 0.9, the IC remains labeled as SOZ in the reference list. As a result, we obtain a comprehensive reference list that categorizes all ICs as either Noise, RSN, or SOZ. All computations were executed on an Intel(R) Core(TM) i7-4790 CPU @ 3.60 GHz with 32GB RAM, running a 64-bit operating system. 
%The training process of \textit{DeepXSOZ} took approximately 6.4 hours, while the testing phase took only 1.02 seconds.

\section{Experiments and Results}

 Our evaluation has four goals: a) evaluate efficacy of \textit{DeepXSOZ}, its variation across age and sex and compare against state-of-the-art techniques, b) evaluate correlation of \textit{DeepXSOZ} performance with surgical outcomes, c) Expert rules ablation studies, to show relative importance of spatial and temporal expert knowledge on the \textit{DeepXSOZ} performance, and d) evaluate the variation of manual IC sorting effort for different levels of reliance on the automation in \textit{DeepXSOZ}.

\subsection{Comparative Techniques}
\label{sec:Comp}
Based on the related work in Table \ref{LR}, we chose the following three state-of-art comparative techniques.\\
%\subsubsection{CNN Architecture}
\noindent{\bf CNN Architecture:}
The CNN based DL technique is the same as the Step 1 (Section \ref{AA}) of \textit{DeepXSOZ}, except that the output layer has three classes instead of two. We also applied cost-sensitive learning in CNN for fair comparsion to give equal importance to all the classes on gradient updates. \\
%\subsubsection{LS-SVM based SOZ classification}
\noindent{\bf LS-SVM based SOZ classification:}
The technique proposed in Hunyadi et al.\cite{b12} was replicated. rs-fMRI IC's spatial and temporal signal features were extracted. To perform an unbiased comparison with \textit{DeepXSOZ}, we also applied SMOTE to generate ICs with SOZ features and balance the three classes. The features extracted of Noise, RSN and SOZ were then used to train a LS-SVM as described in Hunyadi et al.\cite{b12}. \\
\begin{figure}[t]
\centering

\includegraphics[width=70mm ,height = 57mm]{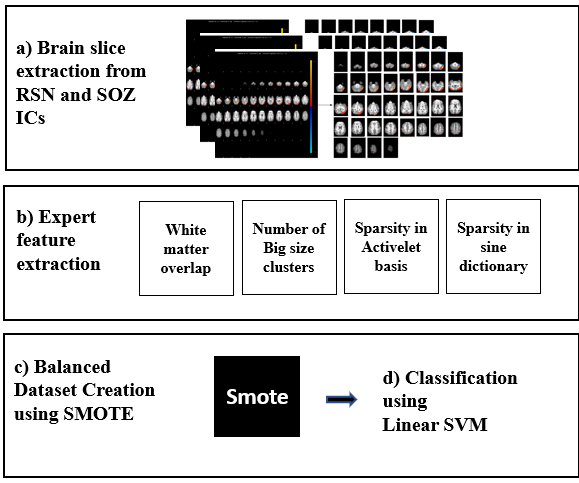}
\caption{Step 2: Feature extraction using expert knowledge.}
\label{fig:Step2}
\end{figure}
%\subsubsection{EPIK: Unsupervised Technique}
\noindent{\bf EPIK:}
EPIK (Banerjee et al.\cite{b44}) uses six expert rules in stage one for an IC to be classified as noise, combined from Boerwinkle and Hunyadi works (Table \ref{LR}). The second stage works on RSN and SOZ classification using their biomarkers.  %presented has the following parts:

\begin{table*}
  \centering
  \scriptsize
  \caption{ { SOZ Identification Performance Metrics. EoK denotes effect of merging expert knowledge in DeepXSOZ.}}
 
  \begin{tabularx}{\textwidth}{@{}>{\bfseries}c*{9}{X}@{}}
  \toprule

  & \textbf{Patient Level Metrics} & \textbf{Method} & \textbf{Age 0-5, N=20 (EoK)}  & \textbf{Age 5-13, N=18 (EoK)} & \textbf{Age 13-18, N=14 (EoK)} 
  & \textbf{Male 
N=23 (EoK)
} & \textbf{Female
N=29 (EoK)
}  & \textbf{Overall Results (EoK)}  & \textbf{DeepXSOZ compare p} \\
  \cmidrule(l){1-10}
 
   & SOZ Accuracy    &\textbf{\textit{DeepXSOZ}}     &\textbf{80.0\% (+30\%)}    & \textbf{88.8\% (+50\%)}   & \textbf{85.7\% (+ 42\%)}  &\textbf{91.3\% (+52\%)}      &\textbf{79.3\% (+37\%)}  &\textbf{84.6\% (+38\%)} & NA\\
   \cmidrule(l){3-10}
   &   & CNN     & 50.0\%
     &38.8\% 
   & 42.8\%      &39.1\%    & 41.9\%   & 46.1\% & p $\approx$ 0\\
    \cmidrule(l){3-10}
   &   & EPIK     & 90.0\%
     &72.2\% 
   &64.2 \%      &78.2\%    &72.4\%    &75.0\% & p = 0.08\\

    \cmidrule(l){3-10}
    
   &        & LS-SVM     & 31.5\%
     & 61.1\% 
   & 71.4\%   &60.8\%    &44.8\%   &50.0\% & p $\approx$ 0\\

   \hline
    & SOZ Precision   &\textbf{\textit{DeepXSOZ}}     & \textbf{94.1\% (+3\%)}   & \textbf{100.0\% (0\%)}    &\textbf{85.7\% (+10\%)}       &\textbf{ 95.4\% (+5\%)}      &\textbf{95.8\% (+9\%)}  &\textbf{93.6\% (+5\%)} & NA   \\
      \cmidrule(l){3-10}
   &   & CNN     &90.9\%
     &100.0\% 
   &75.0\%       &90.0\%    &86.6\%   &88.8\% & p = 0.02\\
     \cmidrule(l){3-10}
   &   & EPIK     & 94.7\%
     &100.0\% 
   &75.0 \%      &94.7\%    & 91.3\%  & 92.8\%  & p = 0.2 \\
   \cmidrule(l){3-10}
   &      & LS-SVM     & 85.7\%
     & 100.0\% 
   & 83.3\%      &93.3\%   & 86.6\%  & 89.6\%  & p = 0.03\\

   \hline
 
 \hline
   & SOZ Sensitivity     &\textbf{\textit{DeepXSOZ}}     & \textbf{84.2\% (+31\%)}   & \textbf{88.8\% (+50\%)}   & \textbf{100\% (+50\%)}      &\textbf{95.4\% (+54\%)}     &\textbf{82.1\% (+37\%)}  &\textbf{89.7\% (+41\%)} & NA\\
     \cmidrule(l){3-10}
   &   & CNN     &52.6\%
     &38.8\% 
   &50.0\%       &40.9\%    &44.8\%  &48.9\%  & p $\approx$ 0\\

    \cmidrule(l){3-10}
   &   & EPIK     & 94.7\%
     &72.2\% 
   &75.0\%      &81.8\%    & 77.7\%  & 79.5\%   &  p = 0.1\\
    \cmidrule(l){3-10}
   &      & LS-SVM     & 33.3\%
     & 61.1\% 
   & 83.3\%      &63.6\%   &48.1\%  &53.6\%  & p $\approx$ 0.0008 \\
   \hline
   \end{tabularx}%
   
  \begin{tabularx}{\textwidth}{@{}>{\bfseries}c*{11}{X}@{}}
  \toprule

  & \textbf{IC Level Metrics} & \textbf{Method} & \textbf{Age(0-5), N=20 (SD)}  & \textbf{Age(5-13), N=18 (SD)} & \textbf{Age(13-18), N=14 (SD)} & \textbf{Age (p)}
  & \textbf{Male 
N=23 (SD)
} & \textbf{Female
N=29 (SD) 
} &\textbf{Sex (p)} & \textbf{Overall Results (SD)} &\textbf{DeepXSOZ compare p}\\
  \cmidrule(l){1-12}
  & Machine       &\textbf{\textit{DeepXSOZ}}     & \textbf{21 (11)}   & \textbf{17 (6)}   & \textbf{16 (4)}  &  \textbf{0.7}   &\textbf{16 (5)}     &\textbf{20 (9)} & \textbf{0.1} &\textbf{18 (8)} & NA \\
  
    \cmidrule(l){3-12}
   & Marked    & CNN     &12 (30)
     &9 (18) 
   &9 (17)   &  0.8 &6 (13)    &14 (29) & 0.5 &10 (23)  &  0.02 \\
    \cmidrule(l){3-12}
   & SOZs  & EPIK     &39 (10)
     &45 (9) 
   &48(10)  &  0.5  &44 (10)   & 43 (10) & 0.2 & 43 (10)  & $\approx$ 0\\
 \cmidrule(l){3-12}
   &         & LS-SVM     &6 (6)
     &4 (6) 
   &6 (2)  &  0.7  &6 (4)   &6 (5) & 0.7 &6 (4) & $\approx$ 0\\
   \hline
   & Accuracy      &\textbf{\textit{DeepXSOZ}}     & \textbf{79.7\% (9)}   & \textbf{82.3\% (3)}   & \textbf{83.5\% (5)}  & \textbf{0.3}    &\textbf{82.5\% (5)}     &\textbf{80.9\% (7)} & \textbf{0.19} &\textbf{81.9 \%(7)} & NA\\
   \cmidrule(l){3-12}  
   & (SD)   & CNN     &88.2\%(2)
     &84.6\%(2) 
   &77.4\% (3)  &  0.44  &87.4\% (2)    &84.4\% (2) & 0.7 &86\% (21)  & 0.24 \\
    \cmidrule(l){3-12}
   &   & EPIK     &59.6\% (12)
     &51.5\% (10) 
   &55.5\% (13)   & 0.5 &53.9\% (10)   &57.7\% (12) &  0.4 &56\% (12)  & $\approx$ 0\\
    \cmidrule(l){3-12}
    &       & LS-SVM     &89.3\% (6)
     &89.2\% (4) &90.7\% (5) & 0.53
   &88.6\% (5)      &90.5\% (4) & 0.09  &89.6\% (5) &  0.09 \\
     \hline
   & Specificity      &\textbf{\textit{DeepXSOZ}}     & \textbf{82.8\% (9)}   & \textbf{85.1\% (4)}   & \textbf{85.7\% (5)}  & \textbf{0.3}    &\textbf{86\% (5)}     &\textbf{83\% (8)} & \textbf{0.7} &\textbf{84.7\% (7)} & NA\\
   \cmidrule(l){3-12}  
   & (SD)   & CNN     &91.6\%(2)
     &89.1\%(2) 
   &81.1\% (4)  &  0.08  &92.9\% (2)    &87.1\% (3) &  0.6 &89.7\% (24)  & 0.4 \\
    \cmidrule(l){3-12}
   &   & EPIK     &60.7\% (14)
     &52.4\% (11) 
   &56.8\% (13)   &  0.7 &55\% (12)   &58.9\% (13) &  0.5 &57.2\% (13)  & $\approx$ 0\\
    \cmidrule(l){3-12}
    &       & LS-SVM     &94.1\% (6)
     &93.7\% (5) &94.2\% (4) & 0.6
   &94.6\% (6)      &94.3\% (4) & 0.3 &94\% (5) &  0.03 \\
   \hline
   & Precision       &\textbf{\textit{DeepXSOZ}}     & \textbf{11.1\% (11)}   & \textbf{17.2\% (11)}   & \textbf{13.5\% (8)}  &   \textbf{0.4}  &\textbf{17.3\% (11)}     &\textbf{11.3\% (9)} & 0.07 &\textbf{14\% (11)} & NA\\
   \cmidrule(l){3-12}  
   &(SD)    & CNN     &23.9\%(33)
     &9.8\%(18) 
   &14.5\% (28)  &  0.56  &13.5\% (24)    &12.4\% (23) & 0.6 &16.2\% (27)  & 0.17 \\
    \cmidrule(l){3-12}
   &   & EPIK     &5.8\% (3)
     &5\% (4) 
   &3.4\% (3)   & 0.42 &5.4\% (4)   &4.2\% (4) & 0.5  &4.7\% (4)  & $\approx$ 0\\
    \cmidrule(l){3-12}
    &       & LS-SVM     &8.4\% (14)
     &19.1\% (18) &13.6\% (16) & 0.63
   &18.1\% (18)      &16.4\% (16) & 0.33 &12.9\% (16) & 0.04  \\
   
   \hline
   &Sensitivity       &\textbf{\textit{DeepXSOZ}}     & \textbf{29\% (27)}   & \textbf{39.8\% (23)}   & \textbf{51\% (30)}  &  \textbf{0.4}   &\textbf{41.3\% (24)}     &\textbf{37.7\% (29)} & \textbf{0.1} &\textbf{36.3\% (28)} & NA \\
  
    \cmidrule(l){3-12}
   & (SD)      & CNN     &25.4\% (42)
     &21.3\% (36) 
   &22\% (31)   &  0.8 &14.2\% (25)    &20.5\% (36) & 0.9 &23.1\% (37)  & 0.37 \\
    \cmidrule(l){3-12}
   &   & EPIK     &41.6\% (27)
     &31.5\% (27) 
   &32.6\% (27)   &  0.4 &34.5\% (28)   &31.4\% (26) &  0.1 &32.5\% (27) & $\approx$ 0\\
    \cmidrule(l){3-12}
   &     & LS-SVM     &9.6\% (17)
     &21.1\% (27) &32.4\% (30) & 0.8
   &21.3\% (26)      &17.9\% (25) & 0.4 &18.1\% (26) &  0.009 \\

    \hline
   &F1-score       &\textbf{\textit{DeepXSOZ}}     & \textbf{16.2\% (16)}   & \textbf{24.6\% (13)}   & \textbf{19.1\% (12)}  &  \textbf{0.8}   &\textbf{22.6\% (14)}     &\textbf{16\% (14)} & \textbf{0.5} &\textbf{14.2\% (19)} & NA \\
  
    \cmidrule(l){3-12}
   & (SD)      & CNN     &16.2\% (22)
     &8.4\% (12) 
   &9.3\% (13)   & 0.06  &5\% (12)    &10\% (17) & 0.09 &12.8\% (20)  &  0.03\\
    \cmidrule(l){3-12}
   &   & EPIK     &9.8\% (6)
     &8.5\% (8) 
   &6\% (6)   &  0.77 &9.2\% (7)   &7.1\% (6) & 0.8  &8.1\% (7) & 0.001\\
    \cmidrule(l){3-12}
   &     & LS-SVM     &7.5\% (13)
     &16.9\% (16) &18.4\% (15) & 0.3
   &15.5\% (15)      &12.7\% (16) & 0.12 &12.9\% (15) &  0.02 \\
  
  \bottomrule
  \end{tabularx}%
  \label{tab:tab4}%
  \vspace{-0.1 in}
\end{table*}%

\subsection{Evaluation Metrics}
 To evaluate the accuracy of \textit{DeepXSOZ}, we undertake two fold approach: a) we assess the agreement between \textit{DeepXSOZ} SOZ labelled ICs and surgically targeted SOZ location for each Engel score group for 25 patients with available surgical resection/ablation outcomes in our dataset, and b) for all 52 patients, we validate the accuracy of \textit{DeepXSOZ}'s generated labels by comparing them with the expert's sorted labels. Two classes of metrics used to judge the success for SOZ identification are observed in recent literature (Table \ref{LR}):

\subsubsection {Patient level metrics (PLM)} 
\label{sec:PLM}
These metrics are patient average and have been commonly used in the most recent works including Hunyadi et al.  \cite{b12}, Zhang et al. \cite{b16}, and Lee et al. \cite{b45}. In the case of focal epilepsy, which is characterized by seizures that originate in a specific area of the brain and affect only one part of the body, these metrics suffice.

\subsubsection {IC level metrics (ILM)}
\label{sec:ILM}
These metrics evaluate the surgical team effort in either narrowing from several SOZ-candidate regions to the one selected for surgical targeting or identifying the multi-focal epilepsy, which originates from multiple areas of the brain and involves regions causing seizures. Accuracy, precision, sensitivity, specificity and F1-score are computed using the standard formula. We have also added Machine marked (MM) SOZ metric which denotes the total number of ICs marked as potential SOZ by an automated technique. This number is directly tied to the surgical team's sorting effort.

\subsection{Statistical Methods}
Statistical methods are utilized to derive the significance of: a) the effect of age and sex on the ILM evaluations, and b) the difference in PLM among different algorithms. For the first aim, we utilize a mixed effects model, where the age and sex along with their combined effect as predictors and a random effect on the patient. For the second aim, we utilized a one sided t-test to evaluate statistical significance of the difference between \textit{DeepXSOZ} and other comparative techniques. The $95\%$ confidence $p$ values are provided in Table \ref{tab:tab4}. 

\subsection{Results}
 The results in Table \ref{tab:tab4} show that \textit{DeepXSOZ} outperforms the LS-SVM, CNN and EPIK in all performance metrics. The table shows PLM and ILM across age and sex. For PLM, we show the effect of merging expert knowledge (EoK) with DL in \textit{DeepXSOZ}, which is the difference between \textit{DeepXSOZ} and CNN. The last column provides the statistical significance of the difference between \textit{DeepXSOZ} and other methods. At PLM, the significantly higher sensitivity of \textit{DeepXSOZ} shows that there are very few FNs as compared to LS-SVM, CNN and EPIK which means \textit{DeepXSOZ} is able to detect the correct SOZ ICs for pediatric PRE patients. At an IC level, \textit{DeepXSOZ} outputs on an average 18 machine marked SOZs which are notably less than the prior techniques. Regarding \textbf{multiple ICs identified as SOZ}, an \textbf{expert rule} denotes that the \textbf{more ICs meeting SOZ criteria} which also \textbf{spatially overlap in a given region, the greater the confidence the expert team has in the validity of the overlapping region being the true SOZ}. However, \textit{DeepXSOZ} \textbf{does not employ this inter-IC dependency, since the goal is to independently classify the ICs}. Overall the results suggests that \textit{DeepXSOZ} may save manual sorting effort for the surgical team. \textit{DeepXSOZ} gives an accuracy of 81.9\% and specificity of 84.7\% in SOZ IC classification. The accuracy and specificity of CNN and LS-SVM is inflated due the presence of high TNs (which is tendency of models in imbalanced class dataset) but a closer look at other metrics will show their sub-optimal performance in SOZ recognition. An IC level higher precision indicates presence of SOZ ICs in the machine marked ICs. The ILM of SOZ sensitivity also conveys the improved capability of \textit{DeepXSOZ} in classifying the correct SOZ ICs as compared to the other techniques. \textit{DeepXSOZ}'s F1-score indicates that it performs the best in automatically recognizing the SOZ IC than any other technique. Table \ref{tab:tab4} also shows the \textit{DeepXSOZ}'s comparison with prior techniques for SOZ identification with respect to age and sex of the patients. For PLM, \textit{DeepXSOZ} maintains statistically stable and higher accuracy, precision and sensitivity across all the age groups and sex distribution, whereas LS-SVM and CNN show significant age and sex-based variance. EPIK, which was not only statistically stable but the second-best performer after \textit{DeepXSOZ} at patient level, failed at IC level with only 4.78\% SOZ IC detection capability out of 43 machine marked SOZs. The $p$ values listed in Table \ref{tab:tab4} demonstrate the statistically significant difference between \textit{DeepXSOZ} and either LS-SVM or CNN for PLM. However, statistically there is an insignificant difference between \textit{DeepXSOZ} and EPIK. 
For ILM, age and sex do not significantly affect any of the techniques. However, \textit{DeepXSOZ} demonstrates a statistically significant advantage over EPIK.  
\subsubsection{Surgical team effort reduction estimation}

\textit{DeepXSOZ} reduces the surgical team time-commitment in hand sorting the medical images by nearly 80\%. Out of 100-140 ICs, \textit{DeepXSOZ} outputs approximately 18 potential SOZ ICs, having high likelihood to contain all SOZs for the pre-surgical evaluation. Effective and accurate removal of noise may drastically reduce the time-consuming process of the neurosurgeons for going through all the ICs to locate the SOZ regions. Out of 49 patients, \textit{DeepXSOZ} correctly identified the SOZ ICs for 44 patients, giving an accuracy of 84.61\%. The five patients missed had only two SOZ ICs per patient, possibly indicating an epileptic network with more subtle findings. %The remaining three patients out of 52 total did not have any SOZ ICs present by expert sorting in the dataset.
\begin{table*}
  \centering
  \scriptsize
  \caption{Performance comparison of Methods across surgical procedures and Engel outcomes.}
 
 \begin{tabularx}{\textwidth}{p{0.4 in}|p{0.55 in}p{0.85 in}|p{0.55 in}p{0.85 in}|p{0.4 in}p{0.85 in}|p{0.4 in}p{0.85 in}}
  \toprule
  & \multicolumn{2}{|p{1.4 in}|}{\textbf{ Ablation Procedures  (N=15)}} & \multicolumn{2}{|p{1.4 in}|}{\textbf{ Resection Procedures  (N=7)}}
  & \multicolumn{2}{|p{1.25 in}|}{\textbf{Engel I outcomes (N = 16)}}
    & \multicolumn{2}{|p{1.25 in}}{\textbf{Engel II outcomes (N = 5) }}\\
 \hline   
   \textbf{Approach} & \textbf{Sensitivity} & \textbf{Precision (SD)} &  \textbf{Sensitivity} & \textbf{Precision (SD)}  & \textbf{Sensitivity} & \textbf{Precision (SD)} & \textbf{Sensitivity} & \textbf{Precision (SD)}
  \\
  \hline
    \textbf{\textit{DeepXSOZ}}     & \textbf{93.3\%}          &\textbf{13.7\% (9)}       &  \textbf{85.7\%}         & \textbf{15.3\% (10)} & \textbf{93.7\%} &\textbf{15.6\% (9)} &\textbf{100\%} &\textbf{14\% (3)}\\
 
    LS-SVM     & 66.6\%        & 16.6\% (19)   & 57.1\%       & 15.4\% (16) &  56.2\% &16.3 (19) &80\% &15.3\% (11)    \\
    CNN     & 33.3\%        & 11.7\% (20)   & 42.5\%      & 9.2\% 
   (11) &  43.7\% &10.5 (15)   &60\% &19.3\% (25)  \\
   EPIK     &73.3\%         &7.9\% (6)     & 71.4\%        & 5.9\%  (5) & 75\% &5.9 (5)  &100\% &5.1\% (3) \\
  \bottomrule
  \end{tabularx}%
  \vspace{-0.2 in}
  \label{tab5}%
\end{table*}%
\begin{table}[htbp]
\caption{Knowledge ablation study (Machine marked SOZ (MM SOZ), Precision , Sensitivity .}
\scriptsize
\begin{center}
\begin{tabular}{|c|c|c|c|}
\hline
\textbf{PLM}& \textit{ Accuracy} & \textit{Precision} & \textit{ Sensitivity}   \\ 
    \hline
    \textit{DeepXSOZ} without temporal features & 84.6\%         & 93.6\%         & 89.7\%        \\ 
     \hline
   
    \textit{DeepXSOZ} without activelet sparsity  &  84.6\%         & 93.6\%         & 89.7\%         \\ 
     \hline
   \textit{DeepXSOZ} without sine sparsity &  84.6\%         & 93.6\%         & 89.7\%         \\ 
    \hline
   \textit{DeepXSOZ} without spatial features & 86.5\%        & 93.7\%        & 91.8\%         \\
   \hline
   \textit{DeepXSOZ} without number of clusters & 75\%        & 92.8\%        & 79.5\%         \\
    \hline
    \textit{DeepXSOZ} without white matter overlap & 86.5\%        & 93.7\%        & 91.8\%         \\
    \hline

    \textbf{ILM}& \textit{ MM SOZ} & \textit{Precision} & \textit{ Sensitivity}   \\ 
    \hline
   
    \textit{DeepXSOZ} without temporal features & 19         & 12.6\%         & 38.8\%        \\ 
     \hline
   
    \textit{DeepXSOZ} without activelet sparsity  &  19        & 13.4\%         & 39.2\%         \\ 
     \hline
   \textit{DeepXSOZ} without sine sparsity &  18         & 13.6\%         & 37.8\%         \\ 
    \hline
   \textit{DeepXSOZ} without spatial features & 51        & 6.2\%        & 53.6\%         \\
   \hline
   \textit{DeepXSOZ} without number of clusters & 16        & 13.8\%        & 34.8\%         \\
    \hline
    \textit{DeepXSOZ} without white matter overlap & 51        & 6.2\%        & 53.6\%          \\
    \hline
  \end{tabular}
  \label{tbl:AblationStudy}
\end{center}
\vspace{-0.2 in}
\end{table}
\subsubsection{Performance on subjects undergoing surgery}
From 24 subjects who had surgery to remove regions identified as rs-fMRI SOZ ICs, 16 achieved seizure freedom (Engel I), and 7 experienced significantly reduced postoperative seizure frequency (Engel II). This suggests that the removed regions likely comprised a substantial part of the epileptogenic network. Performance wise \textit{DeepXSOZ} demonstrated a higher sensitivity of 93.33\% for patients undergoing minimally invasive ablation surgery, making it a promising option. For patients undergoing resection, \textit{DeepXSOZ} maintained a consistent sensitivity of 85.71\%, outperforming other techniques that had more FNs. Furthermore, in patients with Engel I outcome, \textit{DeepXSOZ} exhibited a 93\% agreement with expert sorting, reinforcing its suitability for pre-surgical screening.
%24 subjects in total underwent surgery which removed the regions containing the rs-fMRI SOZ ICs. Of these, 16 subjects became seizure free (Engel I) and 7  subjects displayed significantly reduced post operative seizure frequency (Engel II), indicating the region removed was likely a significant proportion of the epileptogenic network. Table \ref{tab5} shows the \textit{DeepXSOZ}, LS-SVM and CNN performance for the SOZ identification on the 24 subjects that underwent surgery. For the patients undergoing ablation surgery, \textit{DeepXSOZ} showed a higher sensitivity of 93.33\%. The results are very encouraging given that ablation is minimally invasive and thus highly preferred over resection. The performance of \textit{DeepXSOZ} is consistent in the patients undergoing resection with sensitivity of 85.71\% whereas other techniques showed poorer performance due to the presence of a high number of FNs. Moreover, \textit{DeepXSOZ} shows 93\% agreement with expert sorting for patients with Engel 1 outcome. This increases confidence in usage of \textit{DeepXSOZ} in pre-surgical screening.

\subsection{DeepXSOZ knowledge ablation studies}
We evaluate the effect of removing expert rules from \textit{DeepXSOZ} on its performance with respect to the PLM and ILM. We also vary the training data size of \textit{DeepXSOZ} from 20\% to 80\% to derive receiver operating characteristics (ROC) curve, which gives us insight on the trade-off between computational overhead, need for training data, level of reliance on automation, and SOZ identification performance. 

\noindent{\bf \textit{DeepXSOZ} without temporal features:} The BOLD signal temporal features are removed one by one from the SL component of \textit{DeepXSOZ}. We create three unique configurations: a) \textit{DeepXSOZ} without activelet domain sparsity, b) \textit{DeepXSOZ} without sine domain sparsity, and c) \textit{DeepXSOZ} without any temporal features. Table \ref{tbl:AblationStudy} reveals no significant impact on PLM, indicating that removing temporal features has limited effect on the classification of patient's ICs with \textit{DeepXSOZ}. However, \textit{\textbf{removing temporal features slightly decreases SOZ precision and sensitivity, leading to a slight increase in MM SOZ and a minimal increase in the surgical team's workload}}.

%shows that there is statistically insignificant effect on PLM, which indicates that removing temporal features has little effect on the resulting classification of patient's ICs with \textit{DeepXSOZ}. \textit{\textbf{Removing temporal features marginally reduces the SOZ precision and sensitivity, and increases the machine marked SOZ resulting in increased (albeit minimal) workload for the surgical team}}.

\noindent{\bf \textit{DeepXSOZ} without spatial features:} The spatial features are removed from the SL component one by one to create three unique configurations: a) \textit{DeepXSOZ} without number of clusters, b) \textit{DeepXSOZ} without white matter overlap, and c) \textit{DeepXSOZ} without spatial features. At PLM,  slight improvement was observed in all three metrics for both, \textit{DeepXSOZ} without white matter and spatial features. However, these increased metrics were the result of large machine marked ICs presence at ILM, decreased SOZ Incidence and increased sensitivity leading to increased effort by the surgical team. \textit{DeepXSOZ} without incorporating the number of clusters did show minimal decrease in machine marked SOZ ICs but this comes at the cost of reduced SOZ IC sensitivity and PLMs.

\noindent{\bf ROC Curve Analysis:} \textit{DeepXSOZ} has the best PLM performance since the indicated point on its ROC curve is closest to the (0,1) point (Fig. \ref{fig:AblationStudy}). The ILM Sensitivity and manual effort is illustrated using the thin vertical and horizontal solid lines. Removing temporal knowledge has little effect on the ROC, however, ROC drastically becomes poorer when the spatial knowledge is removed. A decrease in IC level specificity (more FPs), results in an increase in manual sorting effort, accompanied by a decrease in the need for training data.

\section{Discussions}
\subsubsection{Efficacy of machine-expert Collaboration for SOZ Identification} 
\begin{figure}
\includegraphics[width=90mm,height=70mm]{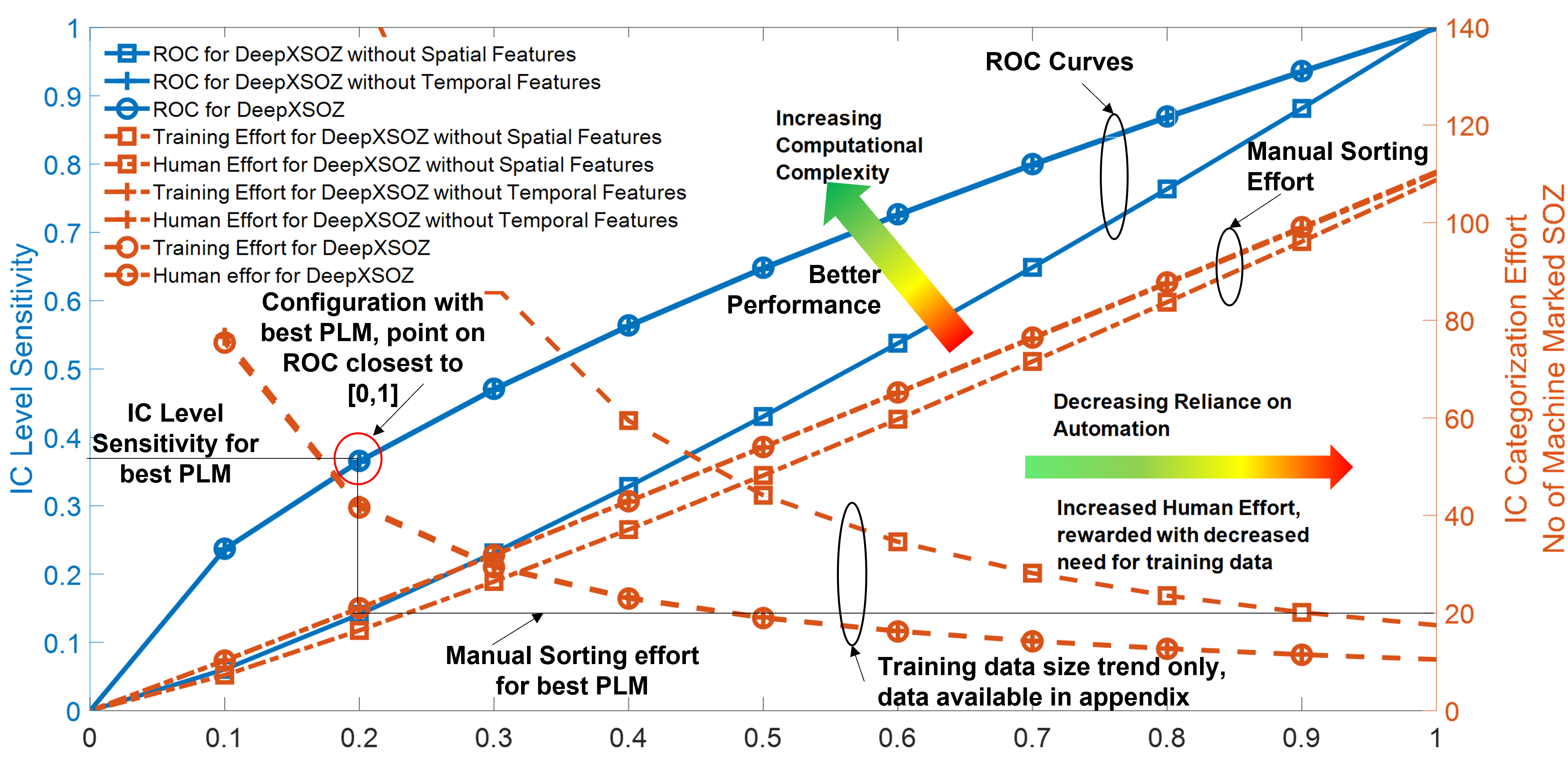}
\caption{Receiver Operating Characteristics and search space for personalization of machine-expert collaboration to identify SOZ.}
\label{fig:AblationStudy}
\vspace{-0.2 in}
\end{figure}

The ROC, manual sorting effort and training data requirement curve in Fig. \ref{fig:AblationStudy} provides a method for personalizing human collaboration with \textit{DeepXSOZ} according to the expertise of the surgical team. The best performing automated technique (chosen configuration red circle) is \textit{DeepXSOZ} that automates spatial and temporal knowledge extraction. However, if the surgical team wants to reduce reliance on automation then a configuration with higher IC level sensitivity can be chosen (moving right along the x axis). This will cause an increase in sorting effort to achieve the same PLM, with a reduction in training data requirement. A surgical team with expertise in manual analysis of spatial features can choose the automation strategy in a solid line with box markers. To reach similar IC level sensitivity as \textit{DeepXSOZ}, it will require more manual effort but less training data. The curve also shows that automation of spatial feature analysis has the most improvements on SOZ identification performance.       

\subsubsection{Clinical Significance of Automated Analysis}
%The rs-fMRI ICA results in approximately 100 ICs. 
In standard rs-fMRI based pre-surgical screening for PRE children, the entire set of ICs is analyzed by a surgical team to determine which ICs capture blood oxygenation changes due to seizure onset. The neurosurgeon then determines the location of SOZ by classifying the patient-level ICs and providing SOZ-candidate locations for iEEG leads for epilepsy surgery evaluation. Given that ICA results in $>$ 100 ICs and only $<$ 10\% are SOZ localizing ICs, manual sorting of rs-fMRI ICs to search for SOZ localizing IC is a significant time commitment and there is low access to centers with available expertise resulting in increased cost and reduced utilization due to low availability, possibly contributing to suboptimal surgical outcomes. An automated whole-brain data-driven SOZ-localizing IC identification technique that is rigorously validated against surgical outcomes, is reproducible, equally effective across age, and sex may greatly improve epilepsy care feasibility and ultimately impact morbidity, and mortality. By reducing the number of ICs to be manually sorted and increasing the confidence of the surgical team on the SOZ localizing capability of an IC, \textit{DeepXSOZ} may potentially alter the number and location of SEEG electrodes, as shown with expert sorting of ICs\cite{b57}, and in certain cases possibly need for sEEG itself. A $>$7-fold  reduction in manual IC processing may expedite diagnostics and enhance implementation feasibility. %Moreover, a $>$7-fold reduction in manual IC processing need may lead to faster diagnostic processes and feasibility of implementation.

\subsubsection{Sedation in rs-fMRI}
Table \ref{tab6} shows that \textit{DeepXSOZ} patient level sensitivity and IC level SOZ precision is statistically unaffected by sedation. \textit{DeepXSOZ}'s sensitivity is significantly higher, statistically stable, and comparable for both sedated and unsedated patients which proves the robustness of this technique. Avoiding conscious sedation can be really helpful as it puts additional risks on the children\cite{b43}. Although this is an exciting result, the number of unsedated patients was only 11 due to the retrospective nature of the study. 
%Prospective study is needed to evaluate automated technique performance with and without sedation.

\begin{table}
  \centering
  \scriptsize
  \caption{Performance of Automated Techniques with and without sedation (N=11 age matched subjects).}
 
 \begin{tabularx}{\columnwidth}{p{0.4 in} p{0.4 in} p{0.75 in} p{0.5 in} p{0.7 in}}
  \toprule
  
   \textbf{Approach} & \textbf{Sensitivity sedation} & \textbf{Precision sedation (SD)} & \textbf{Sensitivity no sedation} & \textbf{Precision no sedation (SD)}
  \\
  \hline
    \textbf{\textit{DeepXSOZ}}             &\textbf{90.9\%}    &\textbf{16\% (10)}  & \textbf{100\%} &\textbf{16.8\% (5.5)}   \\
 
    LS-SVM                &72.7\%  &25.7\% (21.5)   &63.6\% &11.7\%(10.4)\\
    CNN              &45.4 \%  &7.1\% (12.4) &54.5\%    &18.2\% (30)\\
   EPIK             &63.6\%   &5.1\% (4.7) &81.8\%    &3.9\% ( 2.8)   \\
  \bottomrule
  \end{tabularx}%
  \label{tab6}%
  \vspace{-0.2 in}
\end{table}%
%\subsubsection{DeepXSOZ on other datasets}: To adapt \textit{DeepXSOZ} for different datasets coming from disparate institutions, certain fine-tuning may be necessary. Specifically: a) When working with datasets acquired using different MRI machines, adjustments and fine-tuning of \textit{DeepXSOZ} will be required to account for potential variations and optimize performance accordingly, b) If the datasets have been pre-processed using different parameter settings by MELODIC and MCFLIRT, customization of \textit{DeepXSOZ} will be necessary to accommodate these variations and ensure accurate and reliable results, c) The shallow learning component, responsible for feature extraction, will need to be modified in accordance with the characteristics of the newly acquired dataset. This adaptation ensures that \textit{DeepXSOZ} effectively captures the relevant features and maintains optimal performance.
\subsubsection{Limitations and Perspective}
Limitations include a small sample size (n=52) and a single-center design, which may introduce increased variability in \textit{DeepXSOZ}'s performance when applied to larger datasets. While it is one of the largest sets in recent literature for pediatric patients with PRE, a larger study is necessary to address the potential impact of variability in fMRI preprocessing and motion correction techniques, which can differ across centers. Before being used with minimal expert supervision, further testing of \textit{DeepXSOZ} in real-world settings is crucial, considering its intended application in local epilepsy care centers. 
%However, the consistent use of standardized FSL and MELODIC libraries across centers highlights the relevance and reliable performance of \textit{DeepXSOZ} on the new dataset with minor adjustments. 

%\subsubsection{Adult SOZ Identification}
%An intuitive extension of this work is to apply to adult rs-fMRI data. The major difference between seizures in adults and children is that in adults the seizure is most commonly originated from the temporal lope, however, in children there is significantly more heterogeneity in the onset zone. Moreover, movement artifacts in adults may be controlled more easily than in children. This is an important future direction. 

\section{Conclusions}
The most effective treatment for PRE is surgical resection or ablation of the SOZ which requires precise localization to avoid functional brain network damage and developmental impairments. rs-fMRI, a non-invasive technique, can help localize the SOZ and guide iEEG lead placement. However, manual sorting of ICs obtained from rs-fMRI data using ICA is a challenging and subjective task, as only a small fraction (less than 5\%) of the ICs are related to the SOZ. This makes the process time-consuming and limits the reproducibility and availability of this non-invasive technique. To address this, we proposed \textit{DeepXSOZ}, which combines DL with expert knowledge-guided SL to identify SOZ localizing ICs automatically. \textit{DeepXSOZ} identifies ICs of Noise, RSN, and SOZ with significantly improved accuracy compared to previous state-of-the-art techniques. It reduces the number of ICs that need to be analyzed by the surgical team during pre-surgical evaluation, resulting in reduced costs and time. \textit{DeepXSOZ} is compared with other techniques on 52 children with PRE, stratified across age, sex, and one-year post-operative Engel outcomes for rs-fMRI guided resection or ablation. The evaluation is done using PLM and ILM metrics to analyze clinical efficacy and machine-expert collaboration. \textit{DeepXSOZ} achieves significantly higher and consistent results across age, sex, and Engel outcomes in both PLM and ILM from the state-of-the-art solutions and also reduces the number of ICs to be analyzed for pre-surgical evaluation by 7-fold.

%DL techniques on high dimensional data such as biomedical images require a balanced data distribution across the classes of interest. This is often not available for rare diseases where imaging must be performed with utmost care and safeguards against several potential risks. On the other hand, expert features driven image automation techniques tend to provide a high number of FPs. In this paper we demonstrate the integration of DL with expert knowledge through shallow learning from the specific problem domain which  accurately identifies the SOZ in rs-fMRI data for children with PRE. \textit{DeepXSOZ} achieves significantly higher and consistent results across age, sex, and Engel outcomes in both PLM and ILM from the state-of-the-art solutions and also reduces the number of ICs to be analyzed for pre-surgical evaluation by 7-fold. 

\section*{Acknowledgment}
We are thankful to Sarah Wyckoff and Bethany Sussman for rs-fMRI data collection, and subject de-identification.

\end{document}